\documentclass[11pt]{article}

\usepackage[preprint]{acl}

\usepackage{times}
\usepackage{latexsym}

\usepackage[T1]{fontenc}

\usepackage[utf8]{inputenc}

\usepackage{microtype}

\usepackage{inconsolata}

\usepackage{graphicx}

\usepackage{amsmath}
\usepackage{booktabs}
\usepackage{tabularx}
\usepackage{array}
\usepackage{pifont}
\usepackage[table]{xcolor}
\usepackage{multirow}

\usepackage{amsfonts, amssymb} 
\usepackage{float}
\usepackage{xcolor}
\usepackage{tcolorbox}

\usepackage{epigraph}

\usepackage{threeparttable}

\usepackage{soul}

\usepackage{enumitem}
\tcbuselibrary{skins,breakable}
\tcolorboxenvironment{definition}{
  enhanced jigsaw,
  breakable,
  frame hidden,
  colback=blue!2,
  borderline west={1.2pt}{0pt}{blue!45!black},
  sharp corners,
  boxsep=0pt,
  left=6pt,
  right=5pt,
  top=4pt,
  bottom=4pt,
  before skip=6pt,
  after skip=6pt
}

\usepackage{amsthm}
\newtheorem{definition}{Definition}

\usepackage{eso-pic}

\title{A Survey on the Linear Representation \textit{Hypothesis}}

\author{Sewoong Lee\quad Marc E. Canby\quad Ikhyun Cho \quad Julia Hockenmaier \\
  Siebel School of Computing and Data Science \\
  The Grainger College of Engineering \\
  University of Illinois Urbana-Champaign \\
  \texttt{\{samuel27, marcec2, ihcho2, juliahmr\}@illinois.edu}}

\begin{document}

\setcounter{page}{835}
\pagestyle{empty}

\AddToShipoutPictureFG{%
  \setlength{\unitlength}{1mm}%
  \put(105,19){\makebox(0,0){%
    \normalsize\thepage}}%
}

\AddToShipoutPictureFG*{%
  \setlength{\unitlength}{1mm}%
  \put(105,13){\makebox(0,0){\footnotesize
    \textit{Proceedings of the 9th BlackboxNLP Workshop:
    Analyzing and Interpreting Neural Networks for NLP},
    pages 835--851}}%
  \put(105,10){\makebox(0,0){\footnotesize
    October 29, 2026 \copyright 2026
    Association for Computational Linguistics}}%
}

\maketitle
\begin{abstract}
The term ``linear representation hypothesis'' (LRH) has appeared across diverse subfields of artificial intelligence, neuroscience, and cognitive science. But previous works have not consistently treated the LRH as a falsifiable scientific hypothesis; we analyze these inconsistencies and examine their implications for how prior theoretical and methodological results should be interpreted. Based on this analysis, we argue that claims regarding linear representations become well-defined only through careful examination of the model, representation location, feature definition, and evaluation dataset. We therefore propose a more rigorous formalization of the LRH that makes these dependencies explicit and allows the hypothesis to be evaluated as a falsifiable scientific claim. Finally, we identify some non-trivial open problems that warrant further attention from the research community.
\end{abstract}

\section{Introduction}
\label{sec:intro}
Neural activations enable neural systems to do remarkably complex things. Yet it is not obvious how such capabilities arise, which is why the underlying mechanisms of these systems have long been a focus of inquiry.
Neural network activations correspond to $d$-dimensional hidden embedding vectors $\mathrm{emb}(x) \in \mathbb{R}^d$ for inputs $x$.
The discovery of linear regularities in the embeddings of recurrent neural network language models \citep{mikolov2013linguistic}, e.g. 
\(\mathrm{emb}(\text{"king"})-\mathrm{emb}(\text{"queen"})\approx\mathrm{emb}(\text{"man"})-\mathrm{emb}(\text{"woman"})\), likely contributed to the prominence of the concept of linear representations.
The presence of such a relation would suggest the existence of a linear direction in the embedding space such that:
\begin{equation}
\label{eq:m2f}
    \mathrm{emb}(\text{"queen"}) \approx \mathrm{emb}(\text{"king"}) + v_{\text{male}\rightarrow\text{female}},
\end{equation}
where $v_{\text{male}\rightarrow\text{female}} \in \mathbb{R}^d$ is a gender vector (or, more formally, a \textit{counterfactual intervention vector}), providing a more interpretable picture of previously human-incomprehensible hidden embedding vectors.
\begin{table}[t]
\centering
\begin{tabular}{r r r}
\hline
\textbf{Year} & \textbf{New papers} & \textbf{Cumulative} \\
\hline
1976 & 1   & 1 \\
2001 & 1   & 2 \\
2007 & 1   & 3 \\
2022 & 2   & 5 \\
2023 & 11  & 16 \\
2024 & 149 & 165 \\
2025 & 447 & 612 \\
\hline
\end{tabular}
\caption{Number of papers explicitly employing the term ``linear representation hypothesis'' over time, based on Google Scholar search results (counts as of December 2025). The early history is discussed in Section~\ref{sec:history}. The sharp increase after 2022 was triggered by \citet{elhage2022superposition}.}
\label{tab:lrh_over_time}
\end{table}

Whether these early findings generalize across evaluation settings has been subject to debate.
For example, GloVe word embeddings \citep{pennington2014glove} were reported to achieve 75\% accuracy on analogy tests of the form ``\textit{a} is to \textit{b} as \textit{c} is to \underline{~~~}?'' where the answer is predicted by finding the embedding with the highest cosine similarity to
\(\mathrm{emb}(\text{b})-\mathrm{emb}(\text{a})+\mathrm{emb}(\text{c})\).
On the larger BATS analogy benchmark introduced by \citet{gladkova2016analogy}, however, GloVe's accuracy fell below 30\%.
This discrepancy shows that whether linear representations are observed in machine-learned embeddings depends on the embedding model, evaluation dataset, and linguistic features under investigation.

More recently, however, there has been a noticeable shift in terminology: as shown in Table~\ref{tab:lrh_over_time}, the term ``linear representation hypothesis (LRH)'' has come into widespread use.
This naturally leads to a deeper question that has received surprisingly little discussion: \textbf{is \textit{“linear representation”} truly something that merits being called a hypothesis?}
And if so, what would falsifying the LRH imply for research that relies on it as a methodological assumption?

We first categorize the different ways in which the LRH is used in the literature. Some works treat it as a theoretical premise about the structure of neural representations, from which other results are derived (Type 7 in Table~\ref{tab:lrh-contexts}). In others, it is examined empirically, with studies presenting supporting evidence, counterexamples, or analyses of when linear representations appear to hold (Types 2--5; see Table~\ref{tab:lrh-contexts}). Still other work relies on the LRH more implicitly, as a methodological assumption that motivates techniques such as representation steering and related interventions (Type 8 in Table~\ref{tab:lrh-contexts}). Our examination leads us to consider whether the LRH, as commonly formulated, satisfies the criteria of a falsifiable scientific hypothesis. Our analysis suggests that claims about linear representations are only well-defined relative to a specific model, representation location, feature definition, and evaluation dataset; we therefore propose a formalization of the LRH that makes these requirements explicit and places prior empirical findings within the scope of a well-defined hypothesis.

\newcommand{\grayb}{\textcolor{lightgray}{\bullet}}

\newcolumntype{H}{>{\raggedright\arraybackslash}p{3.2cm}}
\newcolumntype{L}{>{\raggedright\arraybackslash}p{2.6cm}}
\newcolumntype{R}{>{\centering\arraybackslash}p{1.0cm}}
\newcolumntype{N}{>{\raggedright\arraybackslash}p{2.8cm}}
\newcolumntype{Y}{>{\raggedright\arraybackslash}X}

\begin{table*}[!htbp]
  \centering
  \small
  \renewcommand{\arraystretch}{1.25}
  \begin{tabularx}{\textwidth}{@{} H N R Y L @{}}
    \toprule
    \textbf{Representation}
      & \textbf{Model}
      & \textbf{Layer}
      & \textbf{Feature}
      & \textbf{Source} \\
    \midrule

    \multirow{12}{*}{Linear representation}
      & OthelloGPT
      & \(\grayb\bullet\grayb\)
      & Board state of Othello
      & \citet{nanda-etal-2023-emergent} \\
    \cmidrule(l){2-5}

      & GPT2-XL, GPT-J, OPT, LLaMA3 8B
      & \(\grayb\bullet\grayb\)
      & Negative-to-positive sentiment shift, detoxification
      & \citet{turner2023steering} \\
    \cmidrule(l){2-5}

      & Claude 3 Sonnet
      & \(\circ\bullet\circ\)
      & The Golden Gate Bridge, implementing addition functions in code, security vulnerability, gender bias, sycophancy, deception, harmful behavior
      & \citet{templeton2024scaling} \\
    \cmidrule(l){2-5}

      & GPT2-small, Pythia 1.4B, Pythia 2.8B
      & \(\grayb\bullet\grayb\)
      & Movie review sentiment
      & \citet{hollinsworth-etal-2024-language} \\
    \cmidrule(l){2-5}

      & LLaMA2 7B
      & \(\circ\circ\bullet\)
      & Gender, translation, morphology, plurality, country--capital; part holonym/meronym not observed
      & \citet{park2024} \\
    \cmidrule(l){2-5}

      & LLaMA2 7B
      & \(\circ\circ\bullet\)
      & Gender, translation, morphology, plurality, country--capital
      & \citet{jiang2024origins} \\
    \cmidrule(l){2-5}

      & LLaMA2, Pythia
      & \(\grayb\bullet\grayb\)
      & Latitude/longitude of places, release date of books
      & \citet{gurnee2024language} \\
    \cmidrule(l){2-5}

      & LLaMA2 7B, LLaMA2 13B, LLaMA2 70B
      & \(\bullet\bullet\bullet\)
      & Truthfulness, honesty
      & \citet{nguyen2025toward} \\
    \cmidrule(l){2-5}

      & LLaMA3 8B
      & \(\circ\circ\bullet\)
      & Hypernym/hyponym
      & \citet{park2025the} \\
    \cmidrule(l){2-5}

      & LLaMA2 7b, Mistral 7b, Vicuna 7b
      & \(\circ\bullet\circ\)
      & Political perspective (liberal/conservative)
      & \citet{kim2025linear} \\
    \cmidrule(l){2-5}

      & Qwen 2.5 7B and LLaMA2 13B
      & \(\grayb\bullet\grayb\)
      & Hallucination, persuasion, pessimism, refusal, sycophancy, truthfulness
      & \citet{agarwal2025context} \\
    \cmidrule(l){2-5}

      & GPT-J, OLMO 7B
      & \(\circ\bullet\bullet\)
      & Relations with high co-occurrence frequency (e.g., country-capital)
      & \citet{merullo2025on} \\

    \midrule

    Onion/concentric representation
      & GRU (RNN)
      & N/A
      & Token position
      & \citet{csordas-etal-2024-recurrent} \\
    \midrule

    Circular representation
      & GPT2-small, Mistral 7B
      & \(\circ\bullet\circ\)
      & Days of the week, months of the year
      & \citet{engels2024not, modell2025origins} \\
    \midrule

    \multirow{2}{*}{Spiral representation}
      & GPT-J, Pythia 6.9B, LLaMA3 8B
      & \(\circ\circ\bullet\)
      & Numbers used in two digit addition
      & \citet{kantamneni2025language} \\
    \cmidrule(l){2-5}

      & LLaMA3 70B
      & \(\bullet\bullet\bullet\)
      & Periodic table
      & \citet{lei2025llamas} \\
    \midrule

    Centroid affinity representation
      & DINOv2, GPT2-Large
      & \(\circ\circ\bullet\)
      & Numbers in MNIST, ``an'' token in language
      & \citet{walker2025centroid} \\
    \midrule

    Representation not found
      & GPT-J, OLMO 7B
      & \(\circ\bullet\bullet\)
      & Relations with low co-occurrence frequency (e.g., company-CEO)
      & \citet{merullo2025on} \\

    \bottomrule
  \end{tabularx}

  \caption{\label{tab:diff}
\textbf{Summary of studies providing empirical evidence for, or counterexamples to, the linear representation hypothesis (LRH).}
The \textbf{Representation} column indicates whether the LRH holds in the study; when it does not hold, the observed non-linear representation is specified.
In the \textbf{Layer} column, $\bullet\circ\circ$ denotes the initial (non-contextualized) embedding layer, $\circ\bullet\circ$ intermediate layers, and $\circ\circ\bullet$ the final layer before unembedding.
Gray circles ($\grayb$) indicate that the corresponding layer was also examined but exhibited weaker linear representations.
}
\end{table*}

\section{Formulation of the LRH}
\label{sec:background}

\subsection{Origins of the LRH}
\label{sec:history}

As shown in Table~\ref{tab:lrh_over_time}, the phrase \emph{linear representation hypothesis} has appeared sporadically before the recent rise of its use in machine learning. These earlier occurrences used the phrase in different contexts: \citet{pitz1976use} used it in cognitive science to describe the hypothesis that numerical samples are mentally encoded along an imaginary number line; \citet{zentall2001case} used it in animal cognition to describe linear orderings over stimuli; and \citet{gonzalez2007supervised} used the related phrase \emph{non-linear representation hypothesis} in the context of classification problems that are not linearly separable in input space.

In contemporary machine learning, the term \emph{linear representation hypothesis} was popularized by \citet{elhage2022superposition}. The rapid increase in usage shown in Table~\ref{tab:lrh_over_time} follows their formulation of the following claim:

\begin{definition}[LRH; \citet{elhage2022superposition}'s]
\label{def:elhage}
Neural networks represent input features as directions in activation space.
\end{definition}

However, as \citet{olah2024nextfivehurdles} points out, the notion of a \emph{feature} in this definition is itself non-trivial.
In this survey, we use \emph{feature} broadly to refer to any concept or relation in an input that prior work seeks to measure in internal representations or manipulate by intervening on those representations; examples are summarized in Table~\ref{tab:diff}.

\subsection{Early Empirical Findings}
\label{sec:empirical_findings}

Building on \citet{elhage2022superposition}'s definition, \citet{nanda-etal-2023-emergent} provide early empirical evidence for the LRH by showing that the board state of an OthelloGPT model trained on Othello move sequences can be recovered from its residual stream using supervised linear probes.
Unsupervised methodological progress has also been made, beginning with the findings by \citet{huben2024sparse} that sparse autoencoders (SAEs) with a single-layer (linear) encoder can find interpretable features, such as words beginning with “w”, from large language models like Pythia-70M \citep{biderman2023pythia}.
Notably, \citet{templeton2024scaling} identify the well-known “Golden Gate Bridge” feature in Claude 3 Sonnet \citep{anthropic2024claude} and shows that increasing this feature's activation causally steers the model's generation toward Golden-Gate-Bridge-related outputs.

\subsection{The Strong, the Weak and the Causal}
\label{sec:strong_weak_causal}

\paragraph{The Strong LRH}

Later, \citet{lewissmith2024} points out that the prior definition (Definition~\ref{def:elhage}) is vague, and shows that it can be understood in two distinct ways:
\begin{definition}[LRH; \citet{lewissmith2024}'s]\label{def:smith}
In a given representation space,
\begin{itemize}
    \item[(i)] (\textbf{Strong LRH}) \textbf{All} features used by neural networks are represented by linear directions.
    \item[(ii)] (\textbf{Weak LRH}) \textbf{Some} features used by neural networks are represented as linear directions.
\end{itemize}
\end{definition}

This classification is meaningful in that it makes at least the strong LRH more easily falsifiable.
Under such constraints, several recent studies provide concrete counterexamples to the strong LRH by identifying features whose representations are inherently non-linear. \citet{csordas-etal-2024-recurrent} demonstrate that GRU models encode certain concepts, such as token position, using concentric representations. Similarly, \citet{engels2024not} argue that transformer-based language models represent periodic concepts, such as days of the week, using circular representation. Together, these results suggest that \emph{not} all features used by neural networks are linearly represented.

\paragraph{The Weak LRH}
In Definition~\ref{def:smith}, the weak LRH asserts the existence of some features that are represented as linear directions.
As summarized in Table~\ref{tab:diff}, several studies report empirical evidence for linear representations of particular features.
However, the existence of supporting examples does not entail that the weak LRH is generally true.
This leads researchers to habitually refer to it as a hypothesis, while leaving unclear how long we should continue to call it a hypothesis. More concretely, the difficulty of generalizing evidence for the weak LRH can be understood along two main dimensions.

First, empirical evidence supporting linear representation for one feature is insufficient to justify the claim that another feature is also linearly represented. Moreover, because each study defines a feature with respect to a particular dataset or input, features that appear identical on the surface may in fact differ. For example, for the linguistic feature of gender, \citet{park2024} examine whether masculine nouns such as ``king'' can be transformed into feminine counterparts like ``queen'', whereas \citet{templeton2024scaling} focus on gender bias using sentence-level contexts such as ``I asked the nurse a question, and ...''. Under this approach, researchers are forced to carry out an endless series of validations for the infinitely many fine-grained features that may exist.

{Second, a more serious challenge is that the class of embedding spaces under investigation is not fixed: they change as new LLMs are developed.} This makes hypothesis formulation in this field particularly challenging. For instance, in physics, while researchers gather evidence in support of a hypothesis, the universe has not suddenly changed into a different one overnight.
By contrast, an LLM under study may eventually become outdated, and researchers will then encounter the embedding spaces of new LLMs.
In other words, experiments that support the LRH for one model do not guarantee generalization to another model trained with different data or architectures.

In sum, although the weak LRH may sound weak by name, it is in fact the most difficult to handle; unlike the strong LRH, it is not easily falsified. 
Because its loose formulation can cover nearly all LRH-related work, references to the LRH in practice typically fall into this category, and we follow this categorization in this paper unless otherwise specified. 
At the same time, because of its broad applicability, the merit of calling it a hypothesis is questionable; since this issue lies at the core of this paper, Section~\ref{sec:hypothetical} will focus on this question in greater depth.

\begin{table*}[h]
\centering
\small
\setlength{\tabcolsep}{6pt}
\renewcommand{\arraystretch}{1.2}
\begin{tabular}{p{0.41\linewidth} p{0.53\linewidth}}
\toprule
\textbf{Context of LRH Usage} & \textbf{Papers} (may appear in multiple rows) \\
\midrule

\textbf{Type 1. Theoretical analyses supporting LRH}
& \citet{park2024, jiang2024origins, park2025the, liu2025predict} \\
\specialrule{0.3pt}{1pt}{1pt}

\textbf{Type 2. Empirical evidence: Supporting examples}

& \citet{nanda-etal-2023-emergent, turner2023steering, templeton2024scaling, hollinsworth-etal-2024-language, park2024, jiang2024origins, gurnee2024language, nguyen2025toward, park2025the, kim2025linear, agarwal2025context} \\
\specialrule{0.3pt}{1pt}{1pt}

\textbf{Type 3. Empirical evidence: Counterexamples}
& \citet{csordas-etal-2024-recurrent, engels2024not, modell2025origins, kantamneni2025language, lei2025llamas, walker2025centroid} \\
\specialrule{0.3pt}{1pt}{1pt}

\textbf{Type 4. Empirical analysis of when LRH holds}
& \citet{merullo2025on} \\
\specialrule{0.3pt}{1pt}{1pt}

\textbf{Type 5. Statistical hypothesis testing of LRH}
& \citet{reizinger2025crossentropy, guo2025quantifying} \\
\specialrule{0.3pt}{1pt}{1pt}

\textbf{Type 6. Critical discussions of LRH limitations}
& \citet{marshall2024refusal, wattenberg2024relational, tan2024analysing, canby2025reliable, muraleedharan2025on, JMLR:v26:23-0058, tiblias2025shape} \\
\specialrule{0.3pt}{1pt}{1pt}

\textbf{Type 7. Theoretical assumption$^\dagger$}
& \citet{boix2024towards, lee2025evaluating, hubotter2025specialization, zhu2025fragility, stevinson2025adversarial}\\
\specialrule{0.3pt}{1pt}{1pt}

\textbf{Type 8. Methodological assumption$^\ddagger$}
& \citet{templeton2024scaling, chanin2024a, ayonrinde2024interpretability, heinzerling-inui-2024-monotonic, skean2024does, bhalla2024interpreting, tan2024analysing, makelov2024is, wu2024reply, ayonrinde2024interpretability, karvonen2024measuring, doi:10.1073/pnas.2417182122, li2025fairsteer, li2025geometry, cho-hockenmaier-2025-toward, agarwal2025context, valois2025vision, engels2025decomposing, mencattini2025exploratory, shen2025llm, chenunderstanding, sutter2025the, lee2025evaluating, yao2025adaptivek, makelov2025towards, qiu-etal-2025-superpose, tas2024words, 10.5555/3766078.3766499, wang2025thoughtprobe, vompa2025beyond, baek2025towards, tolooshams2025sparse, lei2025representation, costa2025flat, pmlr-v267-kalajdzievski25a, chen2025transferring, paulo2025automatically, dalva2025fluxspace, lu2026relational} \\
\specialrule{0.3pt}{1pt}{1pt}

\end{tabular}
\caption{\textbf{Contexts in which the term ``linear representation hypothesis (LRH)'' is used across the literature.}
A substantial fraction of prior work employs the LRH primarily as a methodological assumption (\textbf{Type 8}) to justify linear analysis tools, such as linear probing, sparse autoencoder (SAE), and steering vector. As argued in Section~\ref{sec:from_slogan_to_statement}, when such assumptions are invoked across different layers or feature definitions, the resulting methodological justification may not necessarily transfer across studies.
}
\label{tab:lrh-contexts}
\begin{tablenotes}
\footnotesize
\item $\dagger$ Theoretical findings derived under the assumption that the LRH is true.
\item $\ddagger$ LRH used to justify linear methods (e.g., linear probing, SAE, steering vector).
\end{tablenotes}

\end{table*}

\paragraph{The Causal LRH}
There is also another line of work investigating linear representations of causality \citep{park2024, jiang2024origins, nguyen2025toward}.
Prior work has often cited \citet{park2024} as a seminal contribution.
Their study starts from the premise that causality can be defined by constructing counterfactual pairs between two opposing factual states, as illustrated in Eq.~(\ref{eq:m2f}).
Given such pairs, they characterize the geometry of linear representations through the following three questions:
\begin{definition}[LRH; \citet{park2024}'s]
\label{def:park}
For all \textbf{counterfactual pairs} of a concept,
\begin{itemize}
    \item[(i)] (\textbf{Subspace LRH}) All pairs belong to a shared subspace.
    \item[(ii)] (\textbf{Measurement LRH}) The concept can be measured using a linear probe.
    \item[(iii)] (\textbf{Intervention LRH}) The concept can be modified by adding a steering vector.
\end{itemize}
\end{definition}

A central theoretical contribution of \citet{park2024} is to show that these three definitions can be unified under an appropriate choice of inner product, which they term the \emph{causal inner product}.
Empirically, they further demonstrate that a range of linguistic features, such as male $\rightarrow$ female (gender), English $\rightarrow$ French (translation), and country $\rightarrow$ capital (holonym/meronym), have linear representations that are mutually orthogonal when the concepts are causally separable (see
Table~\ref{tab:diff}).


However, this counterfactual-pair formulation does not naturally extend to all features studied under the LRH, such as the Golden Gate Bridge feature in \citet{templeton2024scaling}.
These limitations are partially addressed by \citet{park2025the}, which extends the framework to general hierarchical\footnote{
Earlier work on hierarchical semantic relations explored representations designed to capture hierarchical structure, including embeddings that encode asymmetric entailment and hypernymy relations \citep{vendrov2016order}, probabilistic denotational representations \citep{lai-hockenmaier-2017-learning}, box embeddings for richer set relations \citep{vilnis-etal-2018-probabilistic}, and hyperbolic embeddings that capture tree-like hierarchies more efficiently than Euclidean embeddings in low dimensions \citep{nickel2017poincare}.
These works were motivated by the view that standard Euclidean representations, including simple linear representations, are not well suited to capturing such relations.
} \textit{is-a} relations by relaxing the reliance on counterfactual pairs, while retaining the constraint that features should be causally separable.\footnote{
%
It is worth noting, however, that empirical claims that causally separable concept vectors are orthogonal under the causal inner product \citep{park2024, park2024the} were later weakened by the observation that even random concepts can become nearly orthogonal in high-dimensional spaces \citep{7vik2024geometry}.
In response, for hierarchical concepts with set inclusion, such as $\text{mammal} \subset \text{animal}$, \citet{park2025geometrycategoricalhierarchicalconcepts} further showed that $\cos\!\left(v_{\text{mammal}} - v_{\text{animal}}, v_{\text{animal}}\right)$ is close to zero. 
%
However, since $v_{\text{mammal}}^{\top}v_{\text{random}} \approx 0$ in high dimensions, this contrast largely reduces to the trivial observation that $\|v_{\text{random}}\|^{2}$ is nonzero;
\citet{7vik2024intricacies} further showed that whitening can make random vectors nearly orthogonal even without causal separability.
}
Nevertheless, \citet{park2024}, \citet{jiang2024origins}, and \citet{park2025the} all share the common limitation that their theoretical analyses apply exclusively to the last layer of the model.
We discuss this layer-specific limitation in more detail in Section~\ref{sec:target_layer}.

\section{Target Embedding Layers}
\label{sec:target_layer}

\paragraph{Last layer}
\label{sec:last-layer}

\citet{park2024} provide a formalization of the linear representation hypothesis for binary concepts defined with counterfactual word pairs (e.g., male $\rightarrow$ female), showing that such concepts are well-defined linear representations as directions in the representation space.
Further, \citet{jiang2024origins} provide a theoretical analysis demonstrating that architectural choices such as softmax and cross-entropy promote a linear structure in the representations used for next-token prediction. 

However, it is important to note that, in both works \citep{park2024, jiang2024origins}, the model is defined as
\[
P(y \mid x) = \text{softmax}(\text{emb}(x)^\top \text{unemb}(y)),
\]
where $x$ denotes the input sequence (e.g., ``He is the'') and $y$ denotes the next token to predict (e.g., ``king'').
By construction, this formulation applies only to the final layer representation immediately preceding the unembedding linear transformation.
Accordingly, the theoretical guarantees established in these works are restricted to the last layer.
Therefore, using these results to justify claims about intermediate layers $1,~\ldots,~L-1$ in a network with a total of $L$ extends their conclusions beyond the scope of what is formally established.

\paragraph{Intermediate layers}

As an early attempt to support the LRH, \citet{nanda-etal-2023-emergent} report that, when linear probes are trained on the hidden states of a GPT model trained on Othello game records to distinguish between white and black stones, performance is relatively poor (62--75\%). This might naively be interpreted as evidence for a non-linear representation. However, when the same probes are trained to distinguish \emph{my} stone versus \emph{your} stone, linear probing achieves extremely high accuracy (99\%) at the 4th layer of the 7-layer model, highlighting that the choice of features is crucial for determining whether a representation is linearly encoded.

Studies measuring linear probing performance and steering effectiveness for sentiment and detoxification tasks similarly find that these effects peak in the middle layers of the network and then decline toward the final layers \citep{turner2023steering, hollinsworth-etal-2024-language}.
Additionally, information about time and position is also linearly represented, with layer-wise performance plateauing around the halfway point of the network \citep{gurnee2024language}. Linear probing and intervention for political perspectives in LLMs are also observed most effective in middle layers \citep{kim2025linear}, and a broad investigation spanning hallucination, persuasion, pessimism, refusal, sycophancy, and truthfulness reports that, in a 28-layer model, linear probing and steering are most effective at the 15th layer \citep{agarwal2025context}.

Recently, \citet{merullo2025on} provide empirical evidence about when linear relations emerge for $(\text{subject}, \text{relation}, \text{object})$ triplets by using the subject-side hidden state $h^{(l)}_{\text{subject}}$ extracted from the $l$-th intermediate layer, and modeling the relation as:
\[
h^{(L)}_{\text{object}} = W h^{(l)}_{\text{subject}} + b,
~\text{where } l \in \{1, \dots, L-1\}.
\]
Building on the findings of \citet{hernandez2024linearity}\footnote{Earlier work by \citet{hernandez2024linearity} observed that some relations (e.g., country--capital) are well captured by faithful linear relational embeddings, whereas others (e.g., company--CEO) are not, despite both being well learned.}, \citet{merullo2025on} further demonstrate that linear relational embeddings tend to emerge when the corresponding triplets have high co-occurrence frequency in the training data, and fail to do so when the frequency is low.
This work is distinguished from prior studies in that it does not merely add supporting or counterexamples to the LRH, but instead seeks to identify the underlying boundary conditions that determine \emph{when} linear representations arise. 

\section{Gap between Theory and Experiment}
\label{sec:gap}

The existing literature on the LRH can be organized along several dimensions, including the perspective taken on LRH, the models studied, the layers examined, and the features considered, as summarized in Table~\ref{tab:diff}. In addition, the extent to which each work provides theoretical versus empirical support for LRH is summarized in Table~\ref{tab:lrh-contexts}.
Having organized the literature by target embedding type, a natural question arises:
If the \textbf{final layer} is where linear structure is theoretically formalized and promoted \cite{park2024, jiang2024origins, park2025the, liu2025predict}, why do empirical studies consistently report the clearest linear separability \citep{marks2024the}, the lowest probing error \citep{nanda-etal-2023-emergent, li2023inference}, and the most effective linear intervention \citep{hollinsworth-etal-2024-language, li2025fairsteer, shen2025llm, kim2025linear, agarwal2025context} at \textbf{intermediate layers}?
Furthermore, as shown in Table~\ref{tab:diff}, the layers that experimentally exhibited non-linear representations were most often the last layer \citep{kantamneni2025language, lei2025llamas, walker2025centroid}.

\citet{skean2025layer} propose an explanation for why the last layer in LLMs is not always the best for embeddings, and why intermediate layers often provide better representations, based on information theory, geometry, and stability analysis.
However, why \emph{linear} representations are empirically observed more frequently in the intermediate layers still remains an open problem, which we provide further details in Section~\ref{sec:open_problems}.

\section{What exactly is hypothetical in LRH?}
\label{sec:hypothetical}
The term \emph{hypothesis} can be used in at least two distinct senses in the literature: scientific hypotheses, which propose explanatory claims about the world, and statistical hypotheses, which are formal statements used within statistical testing procedures \citep{Alger8432}.
However, regarding the LRH, the literature is inconsistent about whether it should be understood as a statistical or a scientific hypothesis.
In this section, we review both possibilities and, in particular, discuss the modifications required for the LRH to function as a falsifiable scientific hypothesis.

\subsection{LRH as a Statistical Hypothesis}
There is a vein of work that approaches the linear representation hypothesis from the perspective of statistical hypothesis testing.
Treating the LRH as a statistical hypothesis means translating a claim about linear representation into a null hypothesis and an alternative hypothesis, along with a test statistic.

\paragraph{Examples of Statistical Formulations.}
For instance, \citet{guo2025quantifying} applied statistical hypothesis testing, finding that embeddings significantly satisfy a linear representation hypothesis, exhibiting stronger linear alignment and additive compositional generalization with respect to interpretable attributes (e.g., gender, age, and occupation) than expected under randomly permuted attribute–embedding pairings.
Another example is \citet{reizinger2025crossentropy}, who treat binary semantic attributes in ImageNet-X \citep{idrissi2022imagenet} defined by human annotations (e.g., object size or position) as proxies for latent factors, and test whether these attributes can be predicted from the second-to-last layer representations of ResNet-50 \citep{he2016deep} and ViT-B/16 \citep{dosovitskiy2021an} using linear decoders at levels significantly above chance.

\paragraph{Limitations of Statistical Formulations.}
Despite their methodological clarity, formulating the LRH as a statistical hypothesis faces two major limitations.
First, specifying an appropriate null hypothesis requires strong assumptions about how embeddings would be structured \emph{in the absence of linear representations}. In practice, such tests do not allow us to accept the hypothesis that an embedding space is structured by linear representations; for example, they only allow us to reject the hypothesis that the embeddings follow a multivariate Gaussian distribution \citep{li2025geometry}, offering limited insight into whether the LRH itself holds.
Second, these approaches rely on assumptions about the underlying data-generating process, including the distributions of latent features or embeddings, such as von Mises–Fisher models \citep{reizinger2025crossentropy} or Monte Carlo simulation \citep{guo2025quantifying}, which are difficult to justify or not clearly generalizable beyond the given dataset.

Nevertheless, viewing the LRH through the lens of statistical hypothesis testing has the advantage of yielding concrete and well-scoped contributions under clearly defined assumptions. However, work that formulates and evaluates the LRH in hypothesis testing remains relatively rare in the literature.

\subsection{LRH as a Scientific Hypothesis}

One of the fundamental requirements of a scientific hypothesis is falsifiability \citep{popper}; an idea that is formulated in a way that does not allow for falsification cannot be regarded as a scientific hypothesis.

\paragraph{Toward Falsifiability.}
In Section~\ref{sec:strong_weak_causal}, we discussed that the definition implicitly used in the majority of the literature corresponds to the weak LRH (Definition~\ref{def:smith}), and that this formulation is inherently challenging to falsify.
Although a number of counterexamples are shown to be represented in non-linear ways (Table~\ref{tab:diff}), these results do not directly refute the weak LRH, 
since they can always escape refutation through post hoc redefinition of features---``well, that wasn’t one of the \emph{\textbf{some}} features!''
Can the LRH, as it is currently formulated and being \emph{increasingly} used (Table~\ref{tab:lrh_over_time}), be regarded as a scientific hypothesis?

The preceding discussion shows that the question \emph{``Is the linear representation hypothesis true or not?''} is not, by itself, a well-defined open question. As Table~\ref{tab:diff} shows, the LRH may hold or fail depending on the model $M$, representation location $l$, feature $f$, and dataset $\mathcal{D}$. The question, therefore, depends on a specific tuple $(M,l,f,\mathcal{D})$. This echoes a fundamental lesson from \citet{gladkova2016analogy} introduced in Section~\ref{sec:intro}: linguistic regularities must be examined together with the data on which they are evaluated.
That is, the LRH must be framed not as a hypothesis about `some' general feature as in previous definitions, but as a hypothesis about `one' specific feature, if it is to function as a falsifiable scientific claim.

\paragraph{Formation vs. Use of Representations.}
A second source of confusion is that the LRH literature often conflates two separate claims. The first is a claim about \emph{representation formation}: whether a feature becomes linearly decodable in a model representation, corresponding to \(x\to \mathrm{emb}(x)\). This is the kind of claim supported by linear probes, as in \citet{nanda-etal-2023-emergent}. 
The second is a claim about \emph{representation use}: whether the downstream computation actually uses that representation when producing the output, corresponding to \(\mathrm{emb}(x)\to y\). This is the kind of claim targeted by causal approaches such as intervention \citep[e.g.,][]{park2024}.

These two claims are logically distinct. As \citet{geiger2021causal} point out, probing is by itself unable to establish that the model causally uses that information in producing its output. 
This distinction also explains why the attempt by \citet{park2024}, discussed in Section~\ref{sec:gap}, to unify different notions of linear representation through the causal inner product is limited as a general definition of the LRH.
Their construction applies a causal inner product to concept vectors obtained from counterfactual token-pair differences in the unembedding space of the final layer. This is essentially different from much of the probing-based literature, where the LRH is often used to determine whether a feature is linearly separable in intermediate-layer representations (For more details, see Appendix~\ref{sec:definitions}).

Consequently, a falsifiable LRH statement must do two things. First, it
must specify the local setting in which the claim is evaluated: the model
\(M\), representation location \(l\), feature \(f\), and evaluation
distribution \(\mathcal D\). Second, it must specify the type of claim
being made: whether the claim concerns representation formation
\((x\to \mathrm{emb}(x))\) or representation use \((\mathrm{emb}(x)\to y)\). 
Therefore, we propose that a falsifiable and clear scientific formulation of the LRH should be stated as follows:
\begin{definition}[LRH; proposed]
\label{def:proposed}
Fix a model \(M\), representation location \(l\), evaluation distribution
\(\mathcal D\) over an input space \(\mathcal X\), and feature specification
\[
\mathcal F=(\mathcal Y_f,y_f,\ell_f,\varepsilon),
\qquad
y_f:\mathcal X\to\mathcal Y_f,
\]
where \(\ell_f\) is the evaluation loss and
\(\varepsilon \ge 0\) its tolerance. Let
\(\mathrm{emb}^{(l)}_M(x)\in\mathbb R^d\) denote the representation of
\(x\) at \(l\).

\begin{itemize}
    \item[(i)] \textbf{Representation formation.}
    \(\mathcal F\) is linearly represented at \(l\) in \(M\) over \(\mathcal D\) iff
    \[
    \begin{aligned}
    &\exists\,W,b: \\[-1mm]
    &
    \mathbb E_{x\sim\mathcal D}\!
    \left[
        \ell_f\!\left(
        W\mathrm{emb}^{(l)}_M(x)\!+\!b,\; y_f(x)
        \right)\!
    \right]\!
    \le \varepsilon .
    \end{aligned}
    \]

\item[(ii)] \textbf{Representation use.}
Given a linear representation satisfying (i), let \(\mathcal I_f\) be a specified intervention on the representation at \(l\).
The representation is \emph{causally used} by \(M\) over \(\mathcal D\) if replacing
\(\mathrm{emb}^{(l)}_M(x)\) with
\(\mathcal I_f(\mathrm{emb}^{(l)}_M(x))\)
produces the corresponding counterfactual change in the model's output
behavior.

\end{itemize}
\end{definition}

Definition~\ref{def:proposed} unifies the different ways the LRH has been
studied in the literature, covering both representation formation and
causal use.
For representation formation, standard multiclass linear probing accuracy
can be expressed within Definition~\ref{def:proposed} by taking \(\mathcal Y_f=[K]\) and writing
\(
z(x)=W\mathrm{emb}^{(l)}_M(x)+b\in\mathbb R^K.
\)
With the \(0\)--\(1\) evaluation loss
\(\ell_f(z,y)=\mathbb I[\arg\max_k z_k\neq y]\),
\[
\mathbb E_{x\sim\mathcal D}
[\ell_f(z(x),y_f(x))]
=
1-\operatorname{Accuracy}_{\mathcal D}(z).
\]
Scalar-valued targets \(y_f(x)\in\mathbb R\) include sparse-autoencoder
feature activations, while vector-valued targets
\(y_f(x)\in\mathbb R^d\) recover the linear-relation setting of
\citet{merullo2025on}, where \(y_f(x)=\mathrm{emb}(o)\).

For representation use, taking \(\mathcal I_f\) to add a concept direction
to the context representation recovers the intervention of
\citet{park2024}, where the intervention changes the target concept in the
model's output while leaving causally separable off-target concepts
unchanged.
Similarly, for
\(h_b=\mathrm{emb}^{(l)}_M(x_{\mathrm{base}})\) and
\(h_s=\mathrm{emb}^{(l)}_M(x_{\mathrm{source}})\),
the intervention used by DAS \citep{geiger2024finding} can be written as
\(
\mathcal I_f(h_b)
=
R^\top\!\left[
(I_d-\Pi)Rh_b+\Pi Rh_s
\right],
\)
where \(R\) is the learned orthogonal rotation and \(\Pi\) is a fixed
orthogonal projector onto a pre-specified subspace of the rotated
representation, whose dimension is chosen in advance as a hyperparameter.

\subsection{From Slogan to Statement} \label{sec:from_slogan_to_statement}
Would there be merit in formulating the hypothesis as in Definition~\ref{def:proposed}?
One reason this question matters becomes clear from Table~\ref{tab:lrh-contexts}: the LRH is frequently used as experimental background for employing specific methods (e.g., linear probing or sparse autoencoder with a single-layer encoder).
However, this practice is a double-edged sword.
On the one hand, building methods on prior work is essential for cumulative scientific progress. 
On the other hand, grounding methodological justification in different layers or unrelated linguistic features can create an illusion that a choice of methods is theoretically justified, even when no shared justification actually exists.
In particular, when it is unclear whether different studies are referring to the same feature and layer, using the LRH as a slogan to motivate linear methodologies leads to claims that are unfalsifiable and thus, borrowing Wolfgang Pauli’s famous phrase, \emph{not even wrong} \citep{10.1098/rsbm.1960.0014}.

Then what is currently built on top of the LRH?
Table~\ref{tab:lrh-contexts} also sheds light on this issue by indicating which lines of work would be affected if the LRH is falsified for a particular combination of model, feature, and layer.
For studies that invoke the LRH as a theoretical assumption (\textbf{Type 7} in Table~\ref{tab:lrh-contexts}), the implications are straightforward: once the LRH is falsified for a given model, feature, or layer, the corresponding theoretical claims no longer hold in that setting.
For studies that rely on the LRH as a methodological justification (\textbf{Type 8} in Table~\ref{tab:lrh-contexts}), while their empirical findings remain valid as observations about specific models and datasets, the methodological justification becomes weaker if the LRH is assumed without being tested.
In particular, when non-linear representations are not explored as alternatives, one cannot rule out the possibility that more appropriate non-linear methods would yield stronger or more informative results \citep{white-etal-2021-non}.

\section{Open problems}
\label{sec:open_problems}
Building on the preceding review of the literature, we highlight two open problems that are well-formulated and warrant further investigation.


\paragraph{The relationship between layer depth and linear representation}
As in Section~\ref{sec:target_layer}, existing research has theoretically examined linear representations of various features and has shown that cross-entropy loss and softmax can promote a linear structure in the final layer of neural networks \citep{park2024, jiang2024origins, park2025the}. However, it remains unclear why empirical studies consistently find stronger linear representations in intermediate layers than at the final layer (Section~\ref{sec:gap}).

\paragraph{Why do linear representations arise only for certain $(M,l,f,\mathcal{D})$ tuples?}
One of the earliest proposed explanations is that neural networks may prefer to retrieve information ``cheaply'' \citep{tom2023worries}, though this remains largely speculative.
Another possibility is that linear representations arise from co-occurrence in pre-training data \citep{merullo2025on}.
However, whether this is the only factor, and whether the relationship extends beyond correlation to causality, remains an open question.
More fundamentally, using Definition~\ref{def:proposed}, we currently lack an explanation of why a feature $f$ exhibits a linear representation for some tuples $(M,l,f,\mathcal{D})$ but not for others.

\section{Conclusion}
\label{sec:conclusion}
In this paper, we have examined the Linear Representation Hypothesis (LRH). We have shown that definitions of the LRH have evolved across prior works and are not always consistent.
In particular, existing studies often draw on results obtained across different layers or feature types.
We propose that any LRH claim should explicitly specify the model, representation location, target feature, and evaluation distribution, as in Definition~\ref{def:proposed}.
Based on a systematic categorization of prior work, we have identified several open problems for future research.

\section*{Limitations}
We deliberately focus on literature that explicitly frames its contributions around the linear representation \emph{hypothesis}.
While this choice provides coherence when organizing an increasingly large body of work, it necessarily excludes many earlier and influential studies that investigate linear structure but do not connect their findings to the LRH.
A notable example is \citet{hewitt-manning-2019-structural}, which demonstrates that distances in syntactic parse trees can be recovered by a linear probe. In that work, linearity is not introduced to support a hypothesis, but rather as a methodological constraint intended to prevent the probe from learning the syntactic parsing task itself.
Such studies fall outside the scope of our survey, although they have inspired later research in this literature.
For a more detailed discussion of representative work in this broader literature and its methodological connections to the LRH, see Appendix~\ref{sec:related-beyond-lrh}.

\section*{Acknowledgments}
We would like to thank Rainer Engelken for introducing us to intriguing issues related to the linear representation hypothesis and for providing inspiration in shaping the direction of this work, and Jinu Lee for helpful feedback on this manuscript.


\bibliography{custom}

\appendix

\section{Connections Between Definitions}
\label{sec:definitions}

\subsection{\citet{park2024}'s Approach}

First, the three different notions of linear representation summarized by \citet{park2024} can be expressed mathematically as follows using the example described in Eq.~\ref{eq:m2f}:

\smallskip

\noindent\textbf{Subspace:}
\[
v_{\text{male}\to\text{female}}
\approx \mathrm{emb}(\text{"queen"})-\mathrm{emb}(\text{"king"}).
\]

\noindent\textbf{Intervention:}
\[
\mathrm{emb}(\text{"king"})+\alpha v_{\text{male}\to\text{female}}
\approx \mathrm{emb}(\text{"queen"}).
\]

\noindent\textbf{Measurement:}\footnote{This can be viewed as a formalization of Definition~\ref{def:elhage}, which states that “neural networks represent input features as directions in activation space.”
Under this formulation, the feature vector can be defined more flexibly than under the other two definitions, since it does not require the feature vector to correspond to a counterfactual concept.
For example, in Section~\ref{sec:empirical_findings}, when using a single-layer sparse autoencoder, each feature is measured as $\mathrm{emb}(x)^{\top} w_{\text{enc}}$, which implicitly adopts this definition as the underlying assumption for feature detection.
Importantly, “being the Golden Gate Bridge” is not itself a counterfactual concept, yet it can still be represented by a measurement of the form $\mathrm{emb}(x)^{\top} v_{\text{GoldenGateBridge}}$.}.

\[
\mathrm{feature}(x)=\mathrm{emb}(x)^\top v_{\text{feature}}.
\]

Here, \citet{park2024} restrict $\mathrm{emb}(x)$ to the final layer where the
softmax is applied in order to derive the following relationship:
\[
\begin{aligned}
P(Y=y\mid x)
&= \mathrm{softmax}\!\left(\mathrm{emb}(x)^{\top}\mathrm{unemb}(y)\right) \\
&= 
\frac{
\exp\!\left(\mathrm{emb}(x)^{\top}\mathrm{unemb}(y)\right)
}{
\sum_{y'} 
\exp\!\left(\mathrm{emb}(x)^{\top}\mathrm{unemb}(y')\right)
},
\end{aligned}
\]
where $Y$ is a random variable representing the next token and $x$ denotes the input text. 
For example, $x=\text{"He is the "}$, $Y(0)=\text{"king"}$, and $Y(1)=\text{"queen"}$.
To measure how much more likely the outcome $Y(1)$ is compared to $Y(0)$, consider the log-odds:
{\small
\[
\log \frac{P(Y(1)\mid x)}{P(Y(0)\mid x)}
= \log
\frac{
\exp(\mathrm{emb}(x)^{\top}\mathrm{unemb}(Y(1)))
}{
\exp(\mathrm{emb}(x)^{\top}\mathrm{unemb}(Y(0)))
}
\]
\[
= \mathrm{emb}(x)^{\top}\mathrm{unemb}(Y(1))
- \mathrm{emb}(x)^{\top}\mathrm{unemb}(Y(0))
\]
\[
= \mathrm{emb}(x)^{\top}
\left(
\mathrm{unemb}(Y(1))-\mathrm{unemb}(Y(0))
\right)
\]
\[
= \mathrm{emb}(x)^{\top} v_{\text{feature}}.
\]
}

While \citet{park2024}'s derivation is advantageous in that $v_{\text{feature}}$ can be easily obtained from next-token unembedding vectors and is more theoretically tractable, the three definitions can be unified without restricting the analysis to the final layer, as will be described in the next section.

\subsection{Beyond the Final Layer}
\label{sec:beyond_the_final_layer}

Formally, if the subspace exists, the definition of intervention follows trivially with $\alpha = 1$.
To see the connection with the measurement definition, suppose we apply an intervention to the embedding of an input text $x$, producing a modified embedding
\(\mathrm{emb}(x)' = \mathrm{emb}(x) + \alpha v_{\text{feature}}.\)
Then,
\[
\mathrm{feature}(\mathrm{emb}(x)')
= (\mathrm{emb}(x) + \alpha v_{\text{feature}})^{\top} v_{\text{feature}}
\]
\[
= \mathrm{emb}(x)^{\top} v_{\text{feature}}
+ \alpha v_{\text{feature}}^{\top} v_{\text{feature}}
\]
\[
= \mathrm{feature}(\mathrm{emb}(x))
+ \alpha \|v_{\text{feature}}\|^2.
\]
Thus, an intervention defined as above leads to a linear change in the feature value measured by the measurement definition.

\subsection{The Fourth Definition}

Linear classifiers and linear separability are foundational concepts in machine learning that long predate the recent LRH literature \citep{minsky_papert_1968}.
In the context of learned representations, linear separability asks whether examples differing in a target feature can be separated by a linear decision boundary in representation space.
This notion is particularly relevant to the LRH because linear probing is widely used to test empirically whether a feature is linearly represented.

\citet{park2024}'s formalization, however, does not cover the standard supervised linear-probing setting, in which the separating direction \(w\) and threshold \(b\) are learned from labeled examples.
This gap motivates considering linear separability alongside the three notions above.

\smallskip
\noindent\textbf{Linear separability:}
\[
\begin{aligned}
\exists\, w,b\quad \text{s.t.:}\quad
w^\top\mathrm{emb}(\text{"queen"}) + b &> 0 \\
w^\top\mathrm{emb}(\text{"king"}) + b &\le 0.
\end{aligned}
\]

Given the measurement definition
\[
\mathrm{feature}(\mathrm{emb}(x))=\mathrm{emb}(x)^\top v_{\text{feature}},
\]
introducing a threshold $b$ yields the linear decision rule
\[
\mathrm{emb}(x)^\top v_{\text{feature}} + b > 0
\]
\[
\quad \text{or} \quad
\mathrm{emb}(x)^\top v_{\text{feature}} + b \le 0.
\]

Thus, adding a bias term \(b\) to \citet{park2024}'s measurement score gives a linear decision rule, with the classifier weight \(w=v_{\text{feature}}\).
This establishes a direct connection to linear separability, but only for the special case in which the separating direction is $v_{\text{feature}}$.
Standard linear probing is more general: given labeled examples, it learns the separating direction \(w\) and threshold \(b\) rather than fixing \(w=v_{\text{feature}}\).
To capture this probing-based setting, the proposed Definition~\ref{def:proposed} can take the same form as a linear probe classifier, $\hat{y}=\sigma(w^{\top}\text{emb}(x)+b)$, where $\sigma(\cdot)$ denotes the sigmoid function.
When the learned probe direction $w$ aligns with $v_{\text{feature}}$, the probe measures the same feature and applies a threshold to determine its presence.

This connection yields a simple corollary: linear probes can be used to test whether a feature is linearly represented.
To encompass the diverse probing methodologies in existing literature, Definition 4 explicitly introduces the $(M, l, f, D)$ tuple. 
For instance, the exact definition of the target feature $f$ can determine whether it appears linearly represented, as shown in recent studies where representations emerge only under specific feature definitions \cite{nanda-etal-2023-emergent}.
Furthermore, modern probes frequently aggregate activations across multiple token positions rather than relying on a single token \cite{kramar2026building}. Definition~\ref{def:proposed} accounts for these variations by requiring the probing location $l$ to be explicitly specified. By formalizing these elements, Definition~\ref{def:proposed} establishes a mapping between empirical linear probes and the theoretical framework of the linear representation hypothesis.

\section{Probing Literature Related to the LRH}
\label{sec:related-beyond-lrh}

This paper mainly focuses on work that explicitly discusses ``the linear
representation hypothesis.'' However, an earlier probing literature
studied closely related questions without using the term. Therefore, this
section provides context for interpreting linear decodability as evidence about linear representations.

\paragraph{Linear probes as diagnostic tools.}
An early influential example is \citet{alain2017understanding}, who use
linear classifiers, which they call ``probes,'' to analyze intermediate
neural representations. By training these probes independently of the
underlying model, they measure how well task-relevant information can
be linearly decoded at different layers. This established linear
probing as a useful diagnostic tool for studying how representations
change across a network.

\citet{hewitt-liang-2019-designing} subsequently questioned how such
probe performance should be interpreted. A high-capacity probe may
learn the probing task itself rather than simply reveal structure
already present in the representation. They therefore introduce
\emph{control tasks} and \emph{selectivity} to distinguish predictive
performance attributable to the representation from that attributable
to probe capacity.

A complementary use of linearity appears in
\citet{hewitt-manning-2019-structural}. Their structural probe learns a
linear transformation under which distances and norms in the
transformed representation recover dependency-tree distances and
depths. Here, linearity serves as a restriction on the probe, limiting
its ability to learn the parsing task independently of the
representation. 
This extends probing beyond categorical prediction to the geometric structure of representations.

\paragraph{Information, accessibility, and complexity.}
\citet{pimentel-etal-2020-information} emphasize that the presence of information in a representation is distinct from its linear decodability. From an
information-theoretic perspective, a more expressive probe may provide
a better estimate of whether information is present at all.
A linear probe therefore addresses the more specific question of whether that information is \emph{linearly decodable}.

This distinction naturally raises the question of probe complexity.
\citet{pimentel-etal-2020-pareto} argue that probe accuracy should be interpreted together with probe complexity: the same predictive performance can provide different evidence depending on how complex a decoder is required to achieve it. 
Similarly,
\citet{voita-titov-2020-information} use minimum description length (MDL) to capture not only final probe performance but also how much effort is required to achieve it, such as the amount of training data needed to reach good predictive performance.
Together, these approaches emphasize that
decodability is not simply binary: the same information may be
accessible with substantially different amounts of data or decoder
complexity.

\paragraph{Beyond linear accessibility.}
Finally, \citet{white-etal-2021-non} show that restricting analysis to linear probes can miss structure that is more naturally recovered non-linearly. Extending the structural-probing framework with a non-linear probe based on an RBF (radial basis function) kernel, they recover syntactic structure more accurately than comparable linear probes across multiple languages. Thus, failure of a linear probe does not necessarily indicate that the target structure is absent; it may simply mean that the structure cannot be recovered well by a linear probe.

Taken together, this literature progressively refines what can be concluded from probing: one must distinguish whether information is present, whether it is linearly decodable, and how much decoder complexity is required to extract it. These considerations provide background for interpreting linear probing as evidence for representation formation under the LRH.


\end{document}